\documentclass{article}
\usepackage{spconf,amsmath,graphicx}
\usepackage{times}
\usepackage{amsmath}
\usepackage{amssymb}
\usepackage{latexsym}
\usepackage{pifont}
\usepackage{makecell}
\usepackage{graphicx}
\usepackage{svg}
\graphicspath{ {./images/} }
\usepackage{url}

\title{GATED MECHANISM FOR ATTENTION BASED MULTIMODAL SENTIMENT ANALYSIS}
\name{Ayush Kumar and Jithendra Vepa}
\address{Observe.AI}
\usepackage{lipsum}
\usepackage{fancyhdr}
\begin{document}
%
\thispagestyle{fancy}%


\maketitle
\begin{abstract}
Multimodal sentiment analysis has recently gained popularity because of its relevance to social media posts, customer service calls and video blogs. 
In this paper, we address three aspects of multimodal sentiment analysis; 1. Cross modal interaction learning, i.e. how multiple modalities contribute to the sentiment, 2. Learning long-term dependencies in multimodal interactions and 3. Fusion of unimodal and cross modal cues. Out of these three, we find that learning cross modal interactions is beneficial for this problem.
We perform experiments on two benchmark datasets, CMU Multimodal Opinion level Sentiment Intensity (CMU-MOSI) and CMU Multimodal Opinion Sentiment and Emotion Intensity (CMU-MOSEI) corpus. Our approach on both these tasks yields accuracies of 83.9\% and 81.1\% respectively, which is 1.6\% and 1.34\% absolute improvement over current state-of-the-art.
\end{abstract}
\section{Introduction}
\label{sec:intro}

Sentiment analysis 
has been one of the widely studied problems in spoken language understanding that aims to determine the opinion of the speaker towards a product, topic or event. With the proliferation of social media platforms such as Facebook, Whatsapp, Instagram and YouTube, huge volume of data is being generated in the forms of podcasts, vlogs, interviews, commentary etc. Multimodal data offer parallel acoustic (vocal expressions like intensity, pitch) and visual cues (facial expressions, gestures) along with the textual information (spoken words), which in particular, provides advanced understanding of affective behavior.



Several approaches have been proposed for multimodal sentiment analysis that attempt to effectively leverage multimodal information.
These are categorised into three types, 1. Methods that learn the modalities independently and fuse the output of modality specific representations \cite{DBLP:journals/expert/WollmerWKSSSM13, poria2017context}, 
2. Methods that jointly learn the interactions between two or three modalities \cite{DBLP:conf/emnlp/ZadehCPCM17, sun2019multi}, and 
3. Methods that explicitly learn contributions from these unimodal and cross modal cues, typically using attention based techniques \cite{DBLP:conf/icmi/ChenWLBZM17, zadeh2018multi, DBLP:conf/acl/MorencyCPLZ18, zadeh2018memory, georgiou2019deep, DBLP:conf/emnlp/GhosalACPEB18}.

\begin{figure*}[ht]
    \centering
    \includegraphics[width=0.75\textwidth]{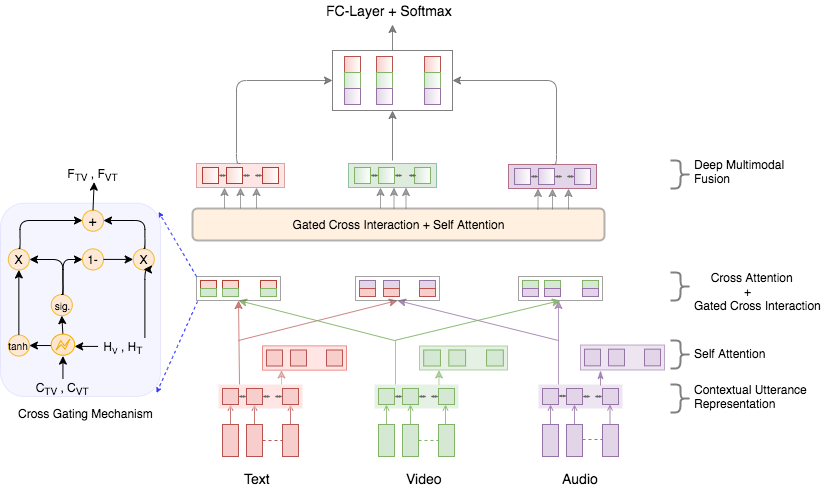}
    \caption{Architectural diagram of the proposed approach.}
    \label{fig:architecture}
\end{figure*}
Most of the existing approaches propose either fusion at different granularities \cite{DBLP:conf/emnlp/ZadehCPCM17, georgiou2019deep} or use a cross interaction block that couple the features from different modalities \cite{DBLP:conf/emnlp/GhosalACPEB18, zadeh2018multi}. 
Combining features from different modalities is necessary as they offer parallel information for same source and help in disambiguation of affective behavior. For example, while uttering sarcastic statements, the speaker shows a distinct intonation which aids in determining the correct sentiment of the speaker.
It is imperative that all modalities in multimodal sentiment analysis do not contribute equally, rather act as cues to reinforce or rectify the information from the other modalities.
This is more evident in the case of imperfect modalities, for example; errors in automatic speech recognition might corrupt the textual information, or poor recording distort the acoustic information, or improper lighting might negatively impact visual features.

Therefore, to learn better cross modal information, we introduce novel \textit{conditional gating mechanism} to modulate the information during cross interactions. Proposed gating mechanism selectively learns the relative importance of different modalities based on the linguistic information, tone of the speaker and facial expressions of an utterance.

Furthermore, to capture long term dependencies across the utterances in the video, we apply a self attention layer on unimodal contextual representations. The major advantage of self attention is that it induces direct interaction between any two utterances and hence offers unrestricted information flow in the network. Finally, we feed the self attended unimodal contextual representations and the gated cross interaction representations to a recurrent layer to obtain \textit{deep multimodal contextual feature vectors} for each utterance.

The main contributions of our proposed approach are: \textbf{1)} Learnable gating mechanism to control information flow during cross interaction, \textbf{2)} Self attended contextual representation to capture long term dependencies, and \textbf{3)} Recurrent layer based fusion of self and gated cross fusion feature vectors to obtain modality specific deep multimodal feature vectors.

\section{Proposed Approach}
\label{sec:approach}
In our proposed model, we aim to learn the interaction between different modalities controlled by learnable gates. Figure \ref{fig:architecture} shows the overall architecture of the system outlining the main components in the model: \textit{contextual utterance representation, self attention, cross attention, gating mechanism for cross interaction} and, \textit{deep multimodal fusion}.

\subsection{Contextual Utterance Representation}
We feed a sequence of utterance level features for each modality to a separate Bi-GRU \cite{DBLP:conf/emnlp/ChoMGBBSB14} and obtain modality specific contextual utterance representation, $H$. Formally, contextual utterance representations ($H_T {\in}R^{u \times d}$) for a sequence of utterances ($U_1, U_2, ..., U_u$) for a \textit{Text} modality can be defined as: 
\begin{equation}
   H_T = Bi\textsf{-}GRU(U_1, U_2, ..., U_u)
\end{equation}
Subscript \textit{T} denotes \textit{Text} modality, \textit{A} and \textit{V} represent \textit{Audio} and \textit{Video} modalities respectively.

\begin{table*}[ht]
\centering
\small
\begin{tabular}{clccc}
\textbf{Sl. No.} & \multicolumn{1}{c}{\textbf{Model}}           & \textbf{CMU-MOSI} & & \textbf{CMU-MOSEI} \\ \hline
B1               & Contextual Unimodal (\textbf{Unimodal Baseline})                             & 80.57      &       & 78.58              \\
B2               & B1 + Self Attention                             & 81.11     &        & 79.12              \\
B3               & Cross Interaction w/o gating (\textbf{Bimodal Baseline})                    & 81.91      &       & 80.00              \\
B4               & Cross Interaction w/ gating                     & 82.91    &         & 80.59              \\
B5               & B2 + B4 w/o deep multimodal fusion              & 83.37   &          & 80.88              \\ \hline
\textbf{B6}      & \textbf{Proposed: B2 + B4 w/ multimodal fusion} & \textbf{83.91} &   & \textbf{81.14}    
\end{tabular}
\caption{Comparison of performance of each step in the proposed model. Accuracy values are mentioned in the table}
\label{tab:baseline}
\end{table*}

\subsection{Self Attention}
In order to capture long term dependencies, a bilinear attention \cite{DBLP:conf/emnlp/LuongPM15} based self matching layer on contextual utterance representations is employed. Since we have sequences of up to 100 utterances in a video, self attention allows us to capture the long context. For a \textit{Text} modality, self attention can be represented as:
\begin{subequations}
\begin{gather}
    M_T = H_TWH_T^{T}, \: M_T \in R^{u \times u}\label{eq:s1}\\
    A_T(i,) = softmax(M_{T_i,})\label{eq:s2}\\
    S_T = A_T.H_T, \: S_T \in R^{u \times d}\label{eq:s3}
    \end{gather}
\end{subequations}
Equation \textit{\ref{eq:s1}} computes the self matching matrix; $W\in R^{d \times d}$ being a trainable matrix, Equation \textit{\ref{eq:s2}} computes self-attention scores for utterance, $U_{i}$ and finally Equation \textit{\ref{eq:s3}} generates the self attended utterance representations.

\subsection{Cross Attention}
Multimodal sentiment analysis provides an opportunity to learn interactions between different modalities. Similar to approaches mentioned for intermodal attention in Ghosal et al  \cite{DBLP:conf/emnlp/GhosalACPEB18}, we propose a method to learn cross-interaction vectors. For a pair of Text ($H_T$) and Video ($H_V$) modalities, co-attention matrix ($M_{TV}{\in}R^{u \times u}$) can be defined as:
\begin{equation}
    M_{TV} = H_TWH_V^{T}; \: W\in{R^{d \times d}}
\end{equation}
Cross attentive representations of Text ($C_{VT}{\in}R^{u \times d}$) and Video ($C_{TV}{\in}R^{u \times d}$) can be represented as:
\begin{subequations}
\begin{gather}
    A_{TV}(i:) = softmax(M_{TV_{i:}})\label{eq:c2}\\
    A_{VT}(:j) = softmax(M_{TV_{:j}})\label{eq:c3}\\
    C_{VT} = A_{VT}.H_T, { } C_{TV} = A_{TV}.H_V\label{eq:c4}
    \end{gather}
\end{subequations}

\subsection{Gating Mechanism for Cross Interaction}
As much as there is an opportunity to leverage cross modal interactions, it brings in challenges of fusing imperfect modalities. To overcome the noise present in individual modalities, we propose a gating mechanism to selectively learn the cross fused vector \cite{DBLP:conf/acl/WangWY18, DBLP:conf/acl/GongB18}. The gated cross fused vector ($F_{PQ} {\in}R^{u \times d}$) for a pair of Text-Video modalities can be obtained as:
\begin{subequations}
\begin{gather}
    F_{VT} = fusion(C_{VT}, H_T) \label{eq:g1}\\
    F_{TV} = fusion(C_{TV}, H_V) \label{eq:g2}
    \end{gather}
\end{subequations}
We define fusion kernel $fusion(\cdot,\cdot)$ 
to be gated combination of cross interaction and contextual representation. Cross interaction, $X(P, Q)$, is a non-linear transformation on cross attended vector ($P$) and contextual representation ($Q$). Gating function, $G(P, Q)$, modulates the information to be passed from cross interaction to next layer. 
\begin{subequations}
\begin{gather}
    X(P,Q) = tanh(W_F.[P, Q, P\textsf{-}Q, P{\circ}Q] + b_F)\label{eq:f1}\\
    G(P,Q) = \sigma(W^T_G.[P, Q, P\textsf{-}Q, P{\circ}Q] + b_G)\label{eq:f2}\\
    F_{PQ} = G(P,Q).X(P,Q) + (1 - G(P,Q)).Q \label{eq:f3}
    \end{gather}
\end{subequations}
where, $W_F$, $b_F$, $W^T_G$, $b_G$ are trainable parameters and ${\circ}$ represents element wise product.

If features from participating modalities are complementary, gating function favours cross interaction and hence would have higher value. On the other hand, if the features from participating modalities is not rich enough or unimodal representation is self-sufficient, the gating function would favor contextual representation and hence would have lower value.

\subsection{Deep Multimodal Fusion}
To aggregate the information from the self and gated cross interactions, we use a Bi-GRU layer to learn deep multimodal feature vector for each modality.
\begin{equation}
    Deep_T = Bi\textsf{-}GRU(S_T, F_{VT}, F_{AT})
\end{equation}

Finally, deep multimodal feature vector for each modality for an utterance is concatenated and fed to the prediction layer containing a fully connected layer followed by softmax layer for final  classification.

\section{Experiments}
\label{sec:experiments}
\subsection{Dataset}
We evaluated our system on two standard multimodal sentiment analysis datasets from CMU multimodal SDK\footnote{ \url{https://github.com/A2Zadeh/CMU-MultimodalSDK}} \cite{zadeh2018multi}, 1) \textit{CMU-MOSI}: CMU Multimodal Opinion level Sentiment Intensity \cite{DBLP:journals/corr/ZadehZPM16} and; 2) \textit{CMU-MOSEI}: CMU Multimodal Opinion Sentiment and Emotion Intensity \cite{DBLP:conf/acl/MorencyCPLZ18}. To compare with the existing approaches, we report results on the binary sentiment classification setup, where \textit{values} $\geq$ 0 signify positive sentiments and \textit{values} $<$ 0 signify negative sentiments. There are 1284, 229 and 686 utterances in the training, validation and the test set for CMU-MOSI dataset while CMU-MOSEI has 16216, 1835 \& 4625 utterances in  training, validation \& test set respectively.

\begin{table*}[t]
\centering
\small
\begin{tabular}{ccc|ccc}
\hline
\multicolumn{3}{c|}{\textbf{CMU-MOSI}}                                                                              & \multicolumn{3}{c}{\textbf{CMU-MOSEI}}                                                           \\ \hline
\textbf{Approach}                                                          & \textbf{Accuracy} & \textbf{F1-Score} & \textbf{Approach}                                        & \textbf{Accuracy} & \textbf{F1-Score} \\
Zadeh et al \cite{DBLP:conf/emnlp/ZadehCPCM17}                         & 77.1              & 79.1              &  Zadeh et al \cite{zadeh2018memory}\textsuperscript{*}      & 76.0              & 76.0             \\
Chen et al \cite{DBLP:conf/icmi/ChenWLBZM17}                          & 76.5              & 73.4              &     Zadeh et al \cite{DBLP:conf/acl/MorencyCPLZ18}       & 76.9              & 77.0   \\
Georgiou et al \cite{georgiou2019deep}                         & 76.9              & 76.9              & Poria et al \cite{poria2017context}               & 77.64              & -              \\
Ghosal et al \cite{DBLP:conf/emnlp/GhosalACPEB18}                       & 82.31             & 80.69             & Ghosal et al \cite{DBLP:conf/emnlp/GhosalACPEB18}     & 79.80             & -                 \\
Sun et al \cite{sun2019multi}                       & 80.6             & 80.57             & Sun et al \cite{sun2019multi}\textsuperscript{\ding{61}}     & (83.62)\textsuperscript{\ding{61}}             & (83.75)\textsuperscript{\ding{61}}                 \\ \hline
\textbf{Proposed Approach}                                              & \textbf{83.91}    & \textbf{81.17}    & \textbf{Proposed Approach}                            & \textbf{81.14 / (85.27)\textsuperscript{\ding{61}}}    & \textbf{78.53 / (84.08)\textsuperscript{\ding{61}}}   
\end{tabular}
\caption{Comparative results on CMU-MOSI and CMU-MOSEI multimodal sentiment analysis. (\textsuperscript{*}) results are taken from Zadeh et al \cite{DBLP:conf/acl/MorencyCPLZ18}, (\textsuperscript{\ding{61}}) results are obtained on CMU-MOSEI dataset after excluding the utterances with sentiment score of 0. We mention the results of proposed model with this setup in the parenthesis.}
\label{tab:benchmarking}
\end{table*}

\begin{table*}[]
\centering
\small
\begin{tabular}{l|c|c|l}
\hline
\multicolumn{1}{c|}{\textbf{Utterance}} & \textbf{\begin{tabular}[c]{@{}c@{}}Gold\\ Label\end{tabular}} & \textbf{\begin{tabular}[c]{@{}c@{}}Predicted\\ Label\end{tabular}} & \multicolumn{1}{c}{\textbf{Remark}} \\ \hline
\textit{I really really loved it}  & Pos. & Pos. & \makecell{$S_{T_u}$ = 0.91, which justifies that text is self sufficient \\cross-interaction score for this utterance is 0.43} \\ \hline
\begin{tabular}[l]{@{}l@{}}\em{i was just thinking about um how its} \\ \em{the performances in it were sort of} \\\em{over overlooked at the academy} \\\em{awards}\end{tabular} & Pos.                                                          & Pos.                                                               & \makecell{Text modality suggests it to be a negative sentiment. Contribu-\\tion of \textit{T-A} and \textit{T-V} cross-interaction is less (0.12 and 0.05). \\$S_{V_u}$ = 0.75 and $S_{A_u}$ = 0.67 suggests that V, A modality \\drives the prediction.} \\ \hline
\textit{maybe only 5 jokes made me laugh}                                                                                                                          & Neg.                                                          & Neg.                                                               & \begin{tabular}[l]{@{}l@{}}\makecell{All three modalities are correlated in this utterance of the video, \\evident by cross-interaction contributions of \textit{T-A}, \textit{A-V} and\\ \textit{T-V} to be 0.78, 0.69 and 0.83 respectively.}\end{tabular}                              \\ \hline
\textit{oh oh my gosh i was blown away}                                                                                                                            & Pos.                                                          & Pos.                                                               & \makecell{Audio ($S_{A_u}$ = 0.62) and video ($S_{V_u}$ = 0.49) contributes in all cross-\\interactions (0.74) to reinforce their learning.}                                                                                                                                                                                                                                         \\ \hline
\end{tabular}
\caption{Qualitative analysis of the proposed model. \textit{T, A, V} refers to text, audio and video respectively. $S_{M_u}$ denotes self attention score for utterance \textit{u} in modality \textit{M}. Cross-interaction score are average values of gate \textit{$G(P, Q)$} for a pair of modalities \textit{P, Q}.}
\label{tab:qualitative}
\end{table*}

\subsection{Implementation Details}
In our experiments, we used same features mentioned in Ghosal et al \cite{DBLP:conf/emnlp/GhosalACPEB18}. Specifically, for CMU-MOSEI dataset, we used Glove embeddings for word features, Facets \footnote{ \url{https://https://pair-code.github.io/facets/}} for visual features and CovaRep \cite{DBLP:conf/icassp/DegottexKDRS14} for acoustic features. For MOSI dataset, we used output of a CNN network for utterance level features, 3D CNN features for visual and openSMILE \cite{eyben2013recent} for acoustic features.

We trained Bi-GRUs with hidden size of $100$ for CMU-MOSI dataset and $200$ for CMU-MOSEI dataset, also used a dropout of 0.4 for regularization and ReLU activation \cite{DBLP:conf/icml/NairH10} in dense layers. We used Adam optimizer \cite{DBLP:journals/corr/KingmaB14} with a learning rate $0.0005$ and a batch size $16$ for CMU-MOSI and $32$ for CMU-MOSEI dataset and, finally train the network for $75$ epochs.



\subsection{Results and Analysis}

\subsubsection{Baselines and Ablation Study}
We carried out several experiments to analyze the contribution of the proposed approach (Table \ref{tab:baseline}). We frame a unimodal (\textbf{B1}) and a bimodal baseline (\textbf{B3}) to compare the impacts of self attention (\textbf{B2}) and gating mechanism (\textbf{B4}). Further we also evaluate the model with deep multimodal fusion (\textbf{B6}). 
We see that by using self attention, the performance of model improves by 0.54\% on MOSI and MOSEI corpora. Gating mechanism improves the accuracy by absolute 1\% on MOSI while multimodal fusion adds additional 0.54\% and 0.26\% accuracy on two corpora. 

The gains in performance over these baselines clearly validates our main hypotheses that attention focused gating selectively learns the noise-robust interactions between different modalities and self attention is required to exploit long term context dependencies present in the video. Finally, deep multimodal feature representations learned using self attended representations and gated cross interactions provides additional gains in the accuracies.

\subsubsection{Benchmarking}
To comprehensively compare our method, we list several baselines for multimodal sentiment analysis. 
Tensor fusion network \cite{DBLP:conf/emnlp/ZadehCPCM17} uses a 3-fold cartesian product on unimodal embeddings;
Context-dependent sentiment analysis \cite{poria2017context} learns context dependent multimodal feature representations;
Memory fusion network (MFN) \cite{zadeh2018memory} proposes a 3-step architecture for multi-view sequential learning using attention network and gated memory; 
Graph-MFN \cite{DBLP:conf/acl/MorencyCPLZ18} replaces attention network in MFN with dynamic fusion graph to learn modal dynamics; 
Gated multimodal-embedding with temporal attention \cite{DBLP:conf/icmi/ChenWLBZM17} performs word level modality fusion using gating; 
Hierarchical fusion \cite{georgiou2019deep} performs 3-step fusion at word, sentence and high level for sentiment classification; 
Deep canonical correlation analysis (DCCA) based multi-modal embeddings \cite{sun2019multi}; and Contextual inter-modal attention based network \cite{DBLP:conf/emnlp/GhosalACPEB18} that proposes a multi-modal attention framework to learn joint-association between multiple modalities \& utterances.

In Table \ref{tab:benchmarking}, we present the comparison of our proposed method with other state-of-the-art approaches. 
Our proposed method outperforms the state-of the-art by 1.6\% (absolute) points for CMU-MOSI corpus and 1.34\% points for CMU-MOSEI corpus. 
Qualitative analysis of our results is presented in Table \ref{tab:qualitative} with a few examples. The analysis demonstrates the effectiveness of the model in selectively attending to the relevant modalities by adjusting the modality specific scores (self attention) as well as cross interactions.

\section{Conclusions and Future Work}
In this paper, we propose an approach to improve the multimodal sentiment analysis using self attention to capture long term context and gating mechanism to selectively learn cross attended features. The gating function emphasize on cross interactions when unimodal information is insufficient to decide the sentiment while it assigns lower weightage to cross modal information when unimodal information is sufficient to predict the sentiment. Evaluations on two well known benchmark datasets (CMU-MOSI and CMU-MOSEI) show that our proposed method is significantly better than the state-of-the-art. In future, we will extend the proposed techniques for real world data, e.g. call center customer conversations, where noise in both Text and Audio modalities is high due to poor audio quality, thus resulting in lower speech recognition accuracies.

\bibliographystyle{IEEEbib}
\bibliography{strings,refs}

\end{document}